\documentclass{article}
\usepackage[preprint]{neurips_2025_custom}
\usepackage{fix-cm}

\definecolor{linkColor}{rgb}{0.2,0.4,0.6}
\usepackage[utf8]{inputenc}
\usepackage[T1]{fontenc}
\usepackage[colorlinks=true,linkcolor=linkColor,citecolor=linkColor,filecolor=linkColor,urlcolor=linkColor]{hyperref}
\usepackage{url}
\usepackage{graphicx}
\usepackage{microtype}

\usepackage{amsmath}
\usepackage{amssymb}
\usepackage{amsfonts}
\usepackage{array}
\usepackage{booktabs}
\usepackage{caption}
\usepackage[inline]{enumitem}
\usepackage{tabularx}

\newcolumntype{L}[1]{>{\raggedright\arraybackslash}p{#1}}
\newcolumntype{Y}{>{\raggedright\arraybackslash}X}

\setlist[itemize]{leftmargin=*, itemsep=2pt, topsep=4pt}
\setlist[enumerate]{leftmargin=*, itemsep=2pt, topsep=4pt}
\newcommand{\IAI}{\mathrm{IAI}}
\newcommand{\IIQ}{\mathrm{IIQ}}
\newcommand{\IIQIndex}{\mathrm{IIQ}_{\mathrm{Index}}}
\newcommand{\MicroIAI}{\mathrm{Micro\ IAI}}
\newcommand{\USD}{\mathrm{USD}}
\newcommand{\MaxExpected}{\mathrm{Max}_{\mathrm{Expected}}}

\title{Intelligence Impact Quotient (IIQ): A Framework for Measuring Organizational AI Impact}
\author{%
Chandan Rajah\thanks{\texttt{chandan.rajah@inceptionai.ai}} \\
Inception, G42
\And
Neha Sengupta\thanks{\texttt{neha.sengupta@inceptionai.ai}} \\
Inception, G42
\And
Federico Castanedo\thanks{\texttt{federico.castanedo@inceptionai.ai}} \\
Inception, G42
\And
Robin Mills\thanks{\texttt{robin.mills@inceptionai.ai}} \\
Inception, G42
\AND
Amit Bahree \\
G42
\And
Ramesh Krishnan Muthukrishnan \\
Inception, G42
\AND
Larry Murray \\
Inception, G42
}
\date{}

\begin{document}

\maketitle

\begin{abstract}
The Intelligence Impact Quotient (IIQ) is a composite metric intended to quantify the depth to which AI systems are integrated into organizational work and their impact. Rather than treating access counts or aggregate token volume as sufficient evidence of impact, IIQ combines a novelty-weighted, time-decayed token stock with usage frequency, a grace-period recency gate, organizational leverage, task complexity, and autonomy. The formulation produces a raw Intelligence Adoption Index (IAI) and a normalized 0--1000 IIQ index for comparison between heterogeneous users and units. We also derive sub-daily update rules and a bounded interpretation layer for estimated efficiency and financial impact. The paper positions IIQ as a deployment-oriented measurement framework: a formal proposal for tracking AI embedding in workflows, not a direct measure of model capability or a substitute for causal productivity evaluation. Synthetic scenarios illustrate how the revised metric distinguishes between frequent low-leverage use, semantically repetitive prompting, and more autonomous, higher-consequence AI-assisted work.
\end{abstract}

\section{Introduction}

Organizations increasingly want to measure whether AI systems are becoming embedded in real work rather than provisioned, piloted, or occasionally accessed. Simple indicators such as seat counts, access logs, or total token volume are useful for monitoring infrastructure usage, but they do not distinguish between superficial experimentation and sustained adoption in consequential workflows. Recent work on the use of AI deployed in the real-world argues similarly for richer descriptions of task composition, workflow integration, and usage patterns in the real world \citep{anthropic-primitives,anthropic-software}.

This paper formalizes the Intelligence Impact Quotient (IIQ), a composite metric designed to measure AI integration at the user and organizational level. IIQ is built around a raw Intelligence Adoption Index (IAI) that combines usage volume with temporal behavior, task distinctness, organizational leverage, task complexity, and autonomy. The objective is not to score the capability of the frontier-model in the abstract. Instead, the objective is to quantify whether AI use is recurring, recent, semantically non-redundant, tied to higher-consequence roles, and associated with more sophisticated modes of work.

The proposed framework is intentionally stateful. Considering adoption as an evolving process rather than a one-time event, allowing older activity to matter less while preserving evidence of sustained integration. At the same time, the formulation is designed to reduce two obvious failure modes of a naive token metric: first, repeated inactivity being penalized multiple times for the same lapse; second, score inflation through replaying near-identical prompts or repeatedly resubmitting large documents.

The contributions of this paper are fourfold:
\begin{enumerate}
\item Defines a \textbf{user-level, multiplicative metric} that combines novelty-weighted token stock, usage frequency, recency, organizational leverage, task complexity, and autonomy.
\item Introduces a \textbf{semantic-distinctness }layer so that repeated prompts contribute less to both effective tokens and frequency mass than genuinely new tasks.
\item Separates \textbf{autonomy from task complexity} and adds a grace-period recency gate so that prolonged inactivity is penalized without stacking a full third penalty onto every short lapse.
\item Provides \textbf{temporal update rules}, a n\textbf{ormalized IIQ index}, and a \textbf{bounded interpretation layer} for efficiency and financial estimates.
\end{enumerate}

The remainder of the paper situates IIQ in related literature, presents the mathematical formulation, describes calibration and interpretation choices, and closes with illustrative scenarios and limitations. That final point matters because evidence on generative-AI productivity suggests that measurable impact may appear first through task reallocation, oversight, and workflow redesign rather than immediate changes in wages or recorded hours \citep{humlum2025still}. IIQ should therefore be read as a structured measurement proposal for deployment behavior, not as a definitive estimate of economic value.

\section{Related Work}

Work on AI impact measurement has begun to move away from simple access counts, license allocations, and aggregate model usage toward more direct descriptions of how AI is used within real workflows. Anthropic's Economic Index introduces a set of ``economic primitives'' for analyg deployed AI use across tasks and occupations \citep{anthropic-primitives}. That agenda is close in spirit to IIQ, which also seeks to move beyond undifferentiated token volume. IIQ differs, however, by defining a user-level stateful metric that combines volume with recency, frequency, semantic distinctness, organizational leverage, task complexity, and autonomy. Anthropic's software-development study further shows that automation-heavy coding-agent workflows differ meaningfully from chatbot-style assistance, even when surface usage volume appears similar \citep{anthropic-software}.

GDPval \citep{patwardhan2025gdpval} is related to IIQ because it evaluates frontier AI models on real-world, economically valuable tasks, rather than on abstract benchmark questions. The benchmark covers 44 occupations across the top nine U.S. GDP-contributing sectors, with tasks based on work from experienced industry professionals. However, GDPval measures model capability, while IIQ measures organizational deployment: whether AI is being adopted, repeated, delegated, and embedded in workflows through novelty-weighted usage, frequency, recency, task complexity, organizational leverage, and autonomy. GDPval can therefore serve as an external reference point for calibrating IIQ’s task-complexity and economic-interpretation layers, helping distinguish what AI systems are capable of doing from whether those capabilities are actually operationalized inside organizations.

A second line of work focuses on autonomy and on what should count as genuinely agentic behavior. Anthropic measures autonomy in practice using large-scale Claude Code and API telemetry, showing that autonomous runs, interruption patterns, and domain vary systematically across real deployments \citep{mccain2026autonomy}. Kasirzadeh and Gabriel argue that AI agents should be characterized along autonomy, efficacy, goal complexity, and generality, rather than treated as a single undifferentiated category \citep{kasirzadeh2025characterizing}. METR complements this perspective by framing agent capability in terms of task horizon: models perform much better on short tasks than on multi-step tasks that would occupy humans for hours \citep{kwa2025longtasks}. Together, these papers motivate IIQ's decision to distinguish routine, analytical, strategic, and agentic work and to track autonomy explicitly rather than assuming it is fully reflected in token count alone.

Related evaluation work also cautions against mapping benchmark performance directly to organizational value. Kapoor et al.\ argue that agent evaluations should consider cost, reproducibility, and robustness, not only raw task accuracy \citep{kapoor2024agents}. This matters for IIQ because the objective here is not to score frontier-model capability in the abstract, but to measure whether AI has become embedded in consequential day-to-day work. In that sense, IIQ is closer to a deployment and utilization metric than to a capability benchmark. Kapoor et al.’s \citep{kapoor2024agents} complements IIQ by arguing that useful agent evaluation must go beyond accuracy to include cost, reproducibility, robustness, and benchmark overfitting. This supports IIQ’s deployment-oriented framing: highly autonomous AI use should count as meaningful organizational integration only when it is reliable, affordable, and reproducible in real workflows.

Finally, empirical labor-market evidence suggests that productivity gains and financial value should be interpreted carefully. Fang et al. \citep{fang2025generative} complement this caution with field-experiment evidence from online retail, showing that GenAI adoption can be linked to causal business outcomes such as sales, conversion rates, and total factor productivity; this supports treating IIQ as a leading impact metric that should ultimately be validated against realized organizational outcomes rather than interpreted as a direct productivity estimate. Humlum and Vestergaard  \citep{humlum2025still} document rapid chatbot adoption and self-reported productivity benefits, but near-zero short-run effects on wages and recorded hours, with change appearing first in task composition, oversight work, and AI-integration activities. This is directly relevant to the interpretation layer proposed later in the paper. IIQ can still serve as a useful leading index of organizational AI impact, but its conversion into hours saved or financial value should be read as a model-based estimate, not as a direct observation of realized labor-market effects.

\section{Method}

\subsection{Problem Definition and Design Goals}

The measurement problem addressed here is to estimate how deeply AI is embedded in an individual's workflow and, by aggregation, within an organizational unit. A useful metric should be sensitive not only to how much AI is used, but also to whether use is sustained, recent, semantically non-redundant, connected to higher-consequence roles, and associated with more autonomous or complex forms of work.

The IIQ framework is therefore designed around five goals:
\begin{enumerate}
\item \textbf{Temporal sensitivity}: older activity should matter less than recent activity.
\item \textbf{Behavioral differentiation}: recurring workflow integration should be distinguished from isolated high-volume prompting.
\item \textbf{Novelty awareness}: repeated or near-duplicate prompts should contribute less than distinct tasks.
\item \textbf{Organizational contextualization}: the same amount of AI output can matter differently depending on the scope of the user's decisions.
\item \textbf{Interpretability}: the metric should be decomposable into understandable components and support downstream normalization or estimation layers.
\end{enumerate}

\subsection{IIQ Formulation}

The base calculation produces the Intelligence Adoption Index. For a given user at time period $p$, the IAI is defined as

\[
\IAI_p = T_p \times F_p \times R_p \times V \times C_p \times A_p
\]

where

\begin{itemize}
\item $T_p$ (\textbf{token stock}) is the time-decayed stock of novelty-weighted tokens.
\item $F_p$ (\textbf{frequency}) is a logarithmic, time-decayed multiplier tracking distinct task engagement.
\item $R_p$ (\textbf{recency}) is a grace-period inactivity gate that activates only after a configurable threshold.
\item $V$ (\textbf{organizational leverage}) is a static multiplier based on the user's scope of influence.
\item $C_p$ (\textbf{complexity}) is a rolling-window average of task complexity over recent distinct work.
\item $A_p$ (\textbf{autonomy}) is a bounded multiplier derived from agent turns and active autonomous runtime.
\end{itemize}

The raw IAI is intentionally unnormalized. It functions as the latent score from which other views, such as a bounded index or financial proxy, can be derived. This separation keeps the core formulation mathematically simple while allowing later calibration choices to vary by deployment.

\subsection{Novelty-Weighted Token Stock}

Naively counting all generated tokens invites a simple gaming strategy: replay the same large prompt repeatedly and accumulate score without demonstrating broader workflow integration. To reduce this failure mode, IIQ first measures the novelty of each interaction relative to recent history.

For an interaction $i$ in period $p$, let $x_{i,p}$ denote a representation of the prompt or task trace, and let $\mathcal{H}_{i,p}$ denote a recent comparison history. Define the novelty weight

\[
\nu_{i,p} = \max\left(0,\ 1 - \max_{j \in \mathcal{H}_{i,p}} \operatorname{Sim}(x_{i,p}, x_j)\right)
\]

where $\operatorname{Sim}(\cdot,\cdot) \in [0,1]$ is a similarity score. In practice, this can be implemented using normalized edit distance, keyword overlap, embedding similarity, or a hybrid of these methods. If no comparable history exists, $\nu_{i,p}$ defaults to 1.

Effective tokens in period $p$ are then

\[
G_p = \sum_i \nu_{i,p} \, t_{i,p}
\]

where $t_{i,p}$ is the token count generated for interaction $i$. The token stock evolves as

\[
T_p = \left(T_{p-1} \times (1 - \alpha_T)\right) + G_p
\]

A daily value of $\alpha_T = 0.05$ corresponds to a half-life of roughly 14 days. Because both $G_p$ and later frequency terms use novelty weights, repeated near-identical prompts contribute substantially less than genuinely distinct work.

\begin{figure}[t]
  \centering
  \includegraphics[width=0.98\linewidth]{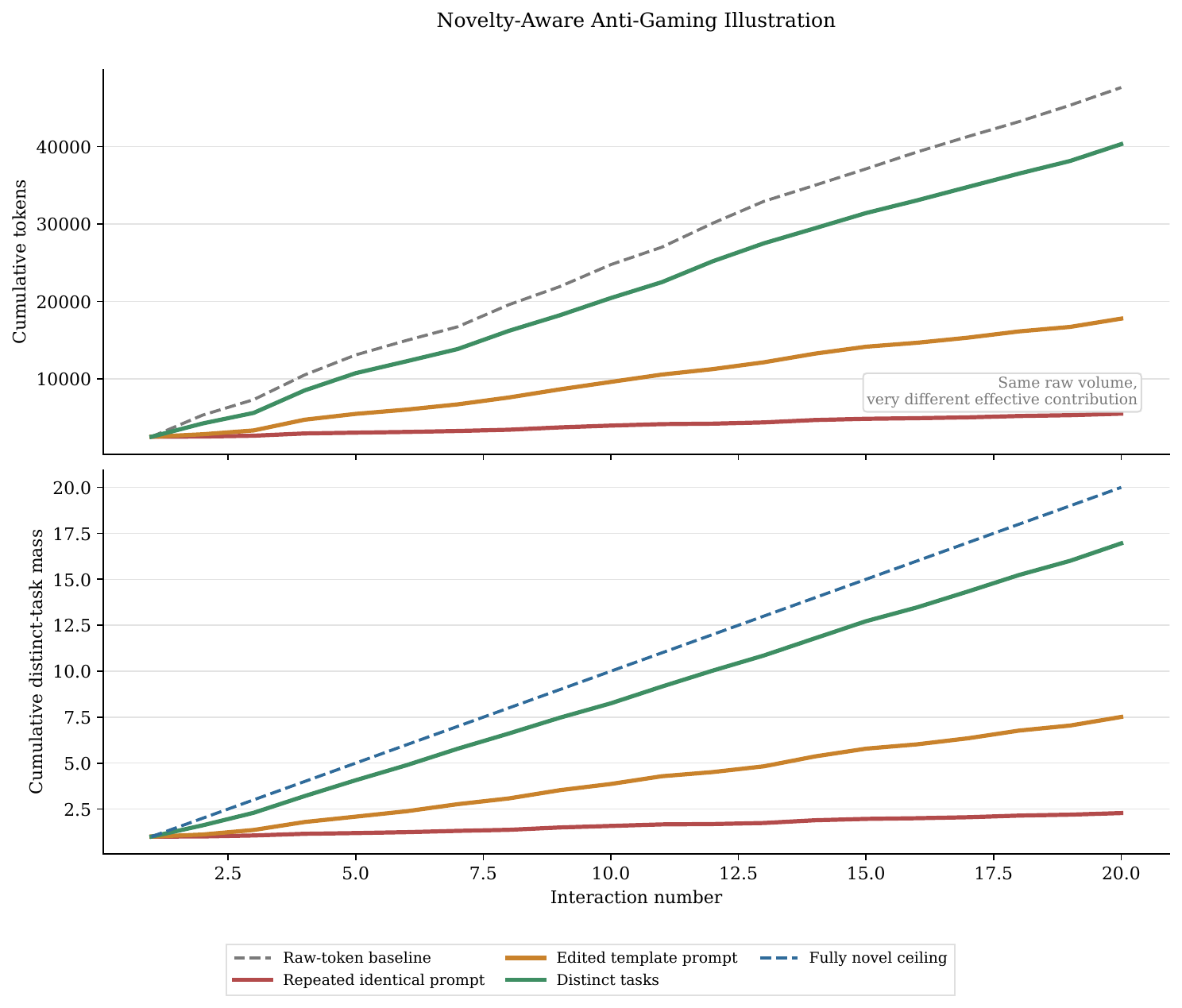}
  \caption{Synthetic illustration of IIQ's novelty mechanism under three 20-interaction usage traces with comparable raw-token volume but different prompt similarity. The top panel compares a shared raw-token baseline against novelty-weighted effective tokens; the bottom panel shows cumulative distinct-task mass, with the dashed ceiling corresponding to a stream in which every interaction is fully novel. The qualitative takeaway is the one the metric is designed to encode: repetitive prompting can keep raw usage high while adding relatively little to measured AI integration, whereas more distinct work increases both the token stock and the frequency signal.}
  \label{fig:novelty-anti-gaming}
\end{figure}

Figure~\ref{fig:novelty-anti-gaming} makes this intuition concrete using synthetic traces with similar raw-token volume but very different prompt similarity patterns. The important point is not that repeated prompting becomes worthless; it is that repetition stops looking like broad workflow integration. A user who repeatedly resubmits the same large prompt still consumes model capacity, but contributes much less to the novelty-weighted token stock and to the distinct-task mass that later drives frequency. By contrast, a user who applies AI across varied tasks accumulates both effective token mass and evidence of recurring integration across different episodes of work.

\subsection{Frequency and Recency}

Frequency measures how often the user engages with distinct tasks independent of total token volume. Let the distinct-task mass in period $p$ be

\[
D_p = \sum_i \nu_{i,p}
\]

The frequency accumulator evolves as

\[
\begin{aligned}
F_{\mathrm{raw},p} &= \left(F_{\mathrm{raw},p-1} \times (1 - \alpha_F)\right) + D_p, \\
F_p &= 1 + \ln\left(1 + F_{\mathrm{raw},p}\right).
\end{aligned}
\]

This construction retains the intuition that repeated use matters, but it discounts semantically repetitive prompting. A user who resubmits the same task many times gains much less frequency mass than a user who uses AI across distinct episodes of work.

Recency remains useful, but here it is designed as a grace-period gate rather than a third continuously compounding decay term. Let $I_p = \operatorname{PeriodsInactive}(p)$ and let $g$ denote the grace threshold. Then

\[
R_p =
\begin{cases}
1, & I_p \le g \\
\exp\left(-\lambda (I_p - g)\right), & I_p > g.
\end{cases}
\]

This means that short gaps are handled primarily by the decay already present in $T_p$ and $F_p$. The recency factor activates only after inactivity extends beyond the configured grace period, making it a churn detector rather than a third full penalty for every brief pause.

\begin{figure}[t]
  \centering
  \includegraphics[width=0.98\linewidth]{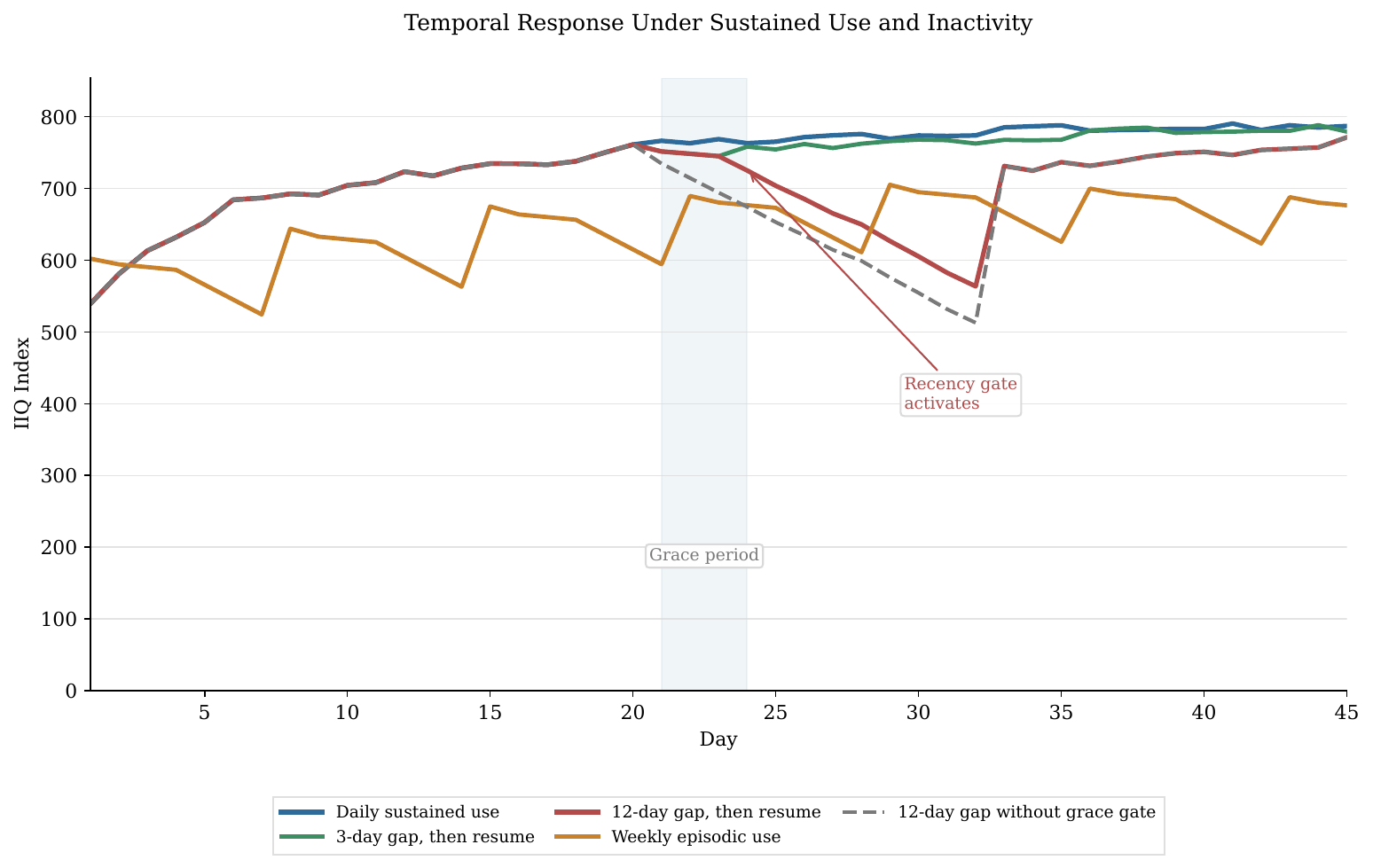}
  \caption{Synthetic 45-day IIQ trajectories for recurring use, a short interruption, a long interruption, and weekly episodic use under the same calibration. Short gaps reduce the score mainly through the decay already present in token stock and frequency, while longer gaps trigger the recency gate and cause a sharper drop; the dashed gray line shows the counterfactual without a grace period. The figure is meant to give an intuitive reading of what the metric says over time: sustained AI integration is sticky, brief lapses are treated as normal, and extended inactivity is interpreted as evidence that the workflow is no longer actively AI-embedded.}
  \label{fig:temporal-response}
\end{figure}

Figure~\ref{fig:temporal-response} visualizes the intended temporal interpretation. Under sustained use, IIQ rises and then stabilizes as token and frequency stocks accumulate against decay. Under a short interruption, the score softens but does not collapse because the existing stocks already encode some memory of prior integration. Under a longer interruption, however, the recency gate activates and the decline steepens. This is important for interpretation: IIQ is not meant to punish every brief pause as churn, but it is meant to distinguish a temporary lull from a meaningful lapse in ongoing AI integration.

\subsection{Organizational Leverage}

The organizational-leverage term captures the idea that the same amount of AI-assisted work can have very different downstream consequences depending on the user's role. AI used by a senior executive to shape strategy, for example, may influence a larger share of organizational decisions than AI used in a highly local task. The same logic extends to public-sector settings in which the scope of influence can extend beyond a single team.

The leverage term is modeled as a discrete mapping based on role tiers. Table~\ref{tab:blast-radius} gives one illustrative calibration.

\begin{table}[t]
\centering
\small
\begin{tabularx}{\linewidth}{L{0.12\linewidth} L{0.28\linewidth} L{0.16\linewidth} Y}
\toprule
Level & Role & Multiplier & Illustrative scope \\
\midrule
1 & Individual Contributor & 1.0 & Baseline \\
2 & Manager / Team Lead & 1.5 & Impacts a team of 5--10 \\
3 & Director & 2.5 & Impacts a department \\
4 & VP / GM & 4.0 & Impacts a business unit \\
5 & C-Suite / Founder & 7.0 & Impacts the entire company \\
6 & Minister / Senior Gov Official & 14.0 & Impacts national policy \\
7 & Head of State & 25.0 & Impacts broad governance \\
8 & Sovereign Ruler & 50.0 & Absolute national and international impact \\
\bottomrule
\end{tabularx}
\caption{Illustrative organizational-leverage tiers.}
\label{tab:blast-radius}
\end{table}

\subsection{Task Complexity}

Task complexity should measure the sophistication of recent work, but it need not decay in the same way as token stock or frequency. Penalizing users simply because they are not continuously inventing harder tasks would make the interpretation unstable. Instead, IIQ treats complexity as a rolling-window average over recent distinct tasks.

Let $\mathcal{W}_p$ denote a window of the last $W$ periods. If $c_{i,\tau}$ is the assigned complexity tier for interaction $i$ in period $\tau$, then

\[
C_p =
\frac{
\sum_{\tau \in \mathcal{W}_p} \sum_i \nu_{i,\tau} \, c_{i,\tau}
}{
\sum_{\tau \in \mathcal{W}_p} \sum_i \nu_{i,\tau}
}
\]

with the convention that $C_p = 1.0$ when the denominator is zero. Table~\ref{tab:complexity-tiers} gives an illustrative task taxonomy.

\begin{table}[t]
\centering
\small
\begin{tabularx}{\linewidth}{L{0.12\linewidth} L{0.2\linewidth} L{0.16\linewidth} Y}
\toprule
Tier & Label & Multiplier & Examples \\
\midrule
1 & Routine & 1.0 & Translations, grammar checks, formatting \\
2 & Analytical & 2.0 & Data synthesis, code generation \\
3 & Strategic & 3.5 & Policy modeling, system design \\
4 & Agentic & 5.0 & Multi-agent orchestration, autonomous execution \\
\bottomrule
\end{tabularx}
\caption{Illustrative task-complexity tiers.}
\label{tab:complexity-tiers}
\end{table}

This definition rewards consistently harder work when it occurs, but it does not impose a mechanical decay penalty simply because a user did not perform an increasingly complex task in every subsequent period.

\subsection{Autonomy}

Task complexity and autonomy are related but distinct. A task can be strategically important while still being executed in a tightly supervised chat loop, and conversely an autonomous multi-step run can reveal a sophisticated use of AI even when the task itself is not the most conceptually difficult. IIQ therefore models autonomy separately.

For interaction $i$ in period $p$, define an autonomy mass

\[
u_{i,p} = \omega_{\tau} \operatorname{AgentTurns}_{i,p} + \omega_h \operatorname{ActiveRunHours}_{i,p}
\]

where $\operatorname{AgentTurns}_{i,p}$ counts model-initiated turns or action steps executed without user intervention, and $\operatorname{ActiveRunHours}_{i,p}$ measures active autonomous runtime rather than idle wall-clock duration. The period-level autonomy mass is

\[
U_p = \sum_i \nu_{i,p} \, u_{i,p}
\]

and the autonomy multiplier is

\[
A_p = 1 + \gamma \ln(1 + U_p).
\]

The logarithm keeps long-running autonomous activity valuable but sublinear, so a 20-hour run is informative without dominating the score indefinitely. Novelty weighting also reduces the benefit of repeatedly launching the same autonomous workflow.

\subsection{Initialization and Boundary Conditions}

To make the recurrence relations well defined, the metric requires explicit base cases. We set

\[
T_0 = 0, \qquad F_{\mathrm{raw},0} = 0, \qquad R_0 = 1, \qquad C_0 = 1, \qquad A_0 = 1.
\]

For a new user with no prior comparable history, novelty weights default to $\nu_{i,p} = 1$ for the first observed interactions. If a complexity or autonomy window is empty, the corresponding multipliers revert to their neutral values. These choices make the metric interpretable for cold-start users and prevent undefined values in sparse histories.

\section{Temporal Updating}

\subsection{Point-in-Time Approximation}

For a single isolated task, it can be useful to define a point-in-time approximation that omits the historical stock variables while retaining novelty and autonomy. For interaction $i$, define

\[
\MicroIAI_i = \nu_i \, t_i \times V \times c_i \times a_i
\]

where $t_i$ is the raw token count, $c_i$ is the task-complexity tier, and $a_i = 1 + \gamma \ln(1 + u_i)$ is the interaction-level autonomy multiplier.

\subsection{Sub-Daily Updates}

To run the metric on a periodic basis, such as every 6 hours, baseline daily decay rates can be converted into period-specific decay rates. If a period $P$ is measured in hours, then $n = 24 / P$ periods occur in a day. The conversion is

\[
\begin{aligned}
\alpha_{\mathrm{periodic}} &= 1 - (1 - \alpha_{\mathrm{daily}})^{P/24}, \\
\lambda_{\mathrm{periodic}} &= \lambda_{\mathrm{daily}} \times (P/24).
\end{aligned}
\]

This preserves the long-run decay behavior while allowing higher-frequency updates.

\subsection{Delta IIQ}

Periodic evaluation also allows a first-difference statistic that captures short-horizon movement in the normalized score:

\[
\Delta \IIQIndex = \IIQ_{\mathrm{Index,current}} - \IIQ_{\mathrm{Index,previous}}
\]

The delta is useful when the objective is to monitor whether adoption is accelerating, stabilizing, or decaying over time.

\section{Calibration and Interpretation}

The raw IAI can be translated into more interpretable derived quantities. These transformations are calibration choices rather than intrinsic properties of the base metric, and they should be tuned to the deployment context.

\subsection{Normalization to the IIQ Index}

Because raw IAI scores can range from near zero to very large values, they can be mapped to a bounded 0--1000 IIQ index using a base-10 logarithmic transformation:

\[
\IIQIndex = \min\left(1000,\ \max\left(0,\ \frac{\log_{10}(\IAI + 1)}{\log_{10}(\MaxExpected)} \times 1000\right)\right)
\]

We set $\MaxExpected = 50{,}000{,}000$ as a high-end reference point corresponding to sustained, high-complexity, autonomy-rich usage by a very high-leverage user. The logarithmic mapping compresses the heavy tail of raw IAI values while preserving ordinal comparisons.

\subsection{Estimating Efficiency Gains}

The interpretation layer for hours saved should depend on current-period activity rather than the cumulative stock $T_p$. Otherwise a long history of usage can imply implausibly large savings in a single week. We therefore define estimated hours saved using effective current-period tokens:

\[
\hat{H}_p = \min\left(\rho \, \operatorname{WorkHoursAvailable}(p),\ \left(\frac{G_p}{1000}\right) \times k \times C_p \times A_p\right)
\]

where $k$ is the baseline hours saved per 1,000 effective tokens, $\rho \in (0,1]$ caps the share of available work time that can be credibly displaced, and $\operatorname{WorkHoursAvailable}(p)$ is the plausible number of work hours in period $p$. For example, if a weekly period contains 40 workable hours and $\rho = 0.75$, then the model cannot credit more than 30 hours saved in that week regardless of token volume.

\subsection{Estimating Financial Impact}

To express that estimate in monetary terms, the model applies a baseline hourly wage $W$. For illustration, we set $W = 40$ USD per hour and interpret organizational leverage as scaling an opportunity-value rate rather than literal payroll savings. The estimated financial impact is then

\[
\USD_p = \hat{H}_p \times (W \times V)
\]

This translation should be understood as a deployment-specific proxy for value at stake, not a direct accounting measure.

\subsection{Assumptions and Scope}

These efficiency and financial translations should be read as interpretive estimates rather than direct observations of realized productivity or economic value. They depend on assumptions about annotation quality, wage baselines, leverage multipliers, prompt-novelty measurement, autonomy logging, and the relationship between generated tokens and task completion. As discussed in Section 2, empirical labor-market studies caution that AI may first reshape task composition and oversight burden before producing visible changes in wages or hours \citep{humlum2025still}. For that reason, the raw IAI and normalized IIQ index should be treated as the primary measurement outputs, with hours and dollar conversions reserved for scenario analysis or deployment-specific calibration.

\section{Illustrative Scenarios}

The following synthetic scenarios, evaluated after a 30-day simulation, illustrate how the multipliers interact in combination. They are not empirical observations; they are visual thought experiments meant to show how different behavioral profiles can arrive at similar or different scores for different reasons.

For completeness, the synthetic traces used here and in Figures~\ref{fig:novelty-anti-gaming} and~\ref{fig:temporal-response} were generated from the same update rules as the main formulation using imagined but plausible hyperparameters rather than hand-drawn outputs. The shared settings are $\alpha_T=\alpha_F=0.05$, a recency grace period of $g=3$ days, $\lambda=0.30$, a 14-period rolling window for complexity, autonomy weights $\omega_{\tau}=0.18$ and $\omega_h=1.6$, and $\gamma=0.18$. Profile generation then varies within realistic synthetic ranges for interactions per active day, log-normal token counts, similarity regimes, complexity mixes, agent turns, and active autonomous runtime. These values are illustrative rather than calibrated, and their role is to produce mechanistic examples that remain faithful to the proposed IIQ update process.

\begin{figure}[t]
  \centering
  \includegraphics[width=0.98\linewidth]{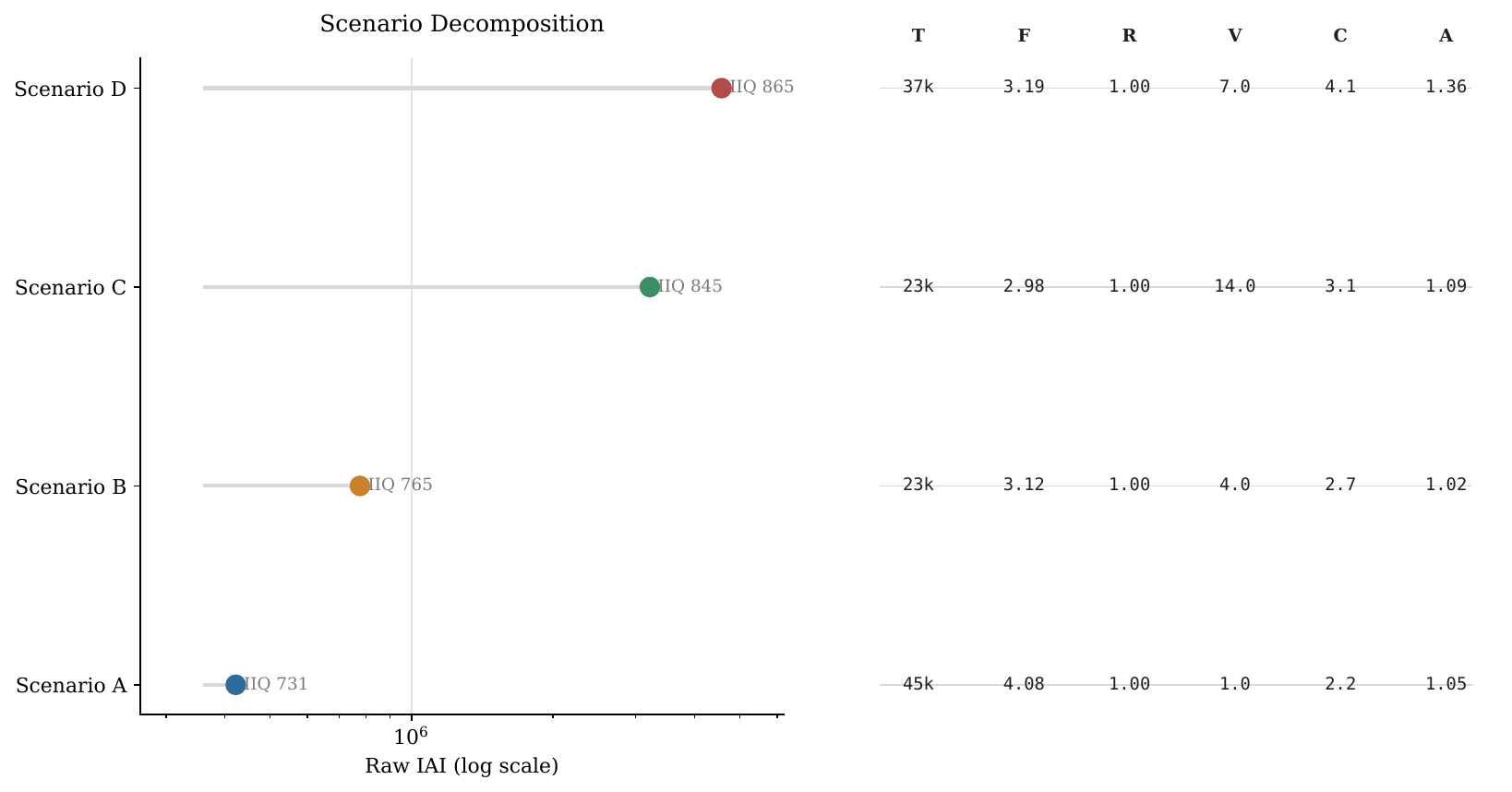}
  \caption{Comparison of four synthetic user profiles after a 30-day simulation. The left panel shows final raw IAI on a log scale, while the right panel lists the realized factor values for token stock $T$, frequency $F$, recency $R$, leverage $V$, complexity $C$, and autonomy $A$. The figure is intended to be read as a decomposition rather than a leaderboard: IIQ is multiplicative, so high-frequency local use, more strategic high-leverage use, and more autonomous episodic workflows can land in similar score ranges through different combinations of factors rather than because any single variable dominates.}
  \label{fig:scenario-decomposition}
\end{figure}

Scenario A is the frequent technical contributor: a user who interacts with AI often, across many distinct tasks, but whose organizational scope is relatively local. The figure shows why this user still scores meaningfully even with low leverage: recurring distinct use steadily builds both token stock and frequency.

Scenario B is a higher-leverage manager or business-unit leader whose use is less frequent but attached to broader decision scope. Here the score is not driven by constant day-to-day prompting. Instead, leverage and moderately higher-complexity work compensate for lower frequency, producing a profile that differs qualitatively from Scenario A even if the final magnitude is not radically different.

Scenario C is a policy or strategy user whose work is less routine, more consequential, and somewhat more autonomous. This profile illustrates the intended effect of combining leverage, complexity, and moderate autonomy: relatively sparse but higher-consequence use can outrank more frequent local usage without requiring extreme token volume.

Scenario D is the episodic but more agentic workflow. The user is not continuously active, so frequency is lower and long pauses remain visible. Even so, the figure shows that complex, autonomy-rich bursts of activity can still push the score upward when they occur in consequential workflows. This is exactly the kind of case for which a purely frequency-based or purely token-based metric would be misleading.

Taken together, the scenarios are meant to give the reader a more intuitive feel for the metric's semantics. IIQ does not say that one canonical pattern of use is always best. It says that recurring integration, non-redundant task variety, organizational scope, task sophistication, and autonomous execution all contribute, and that different usage styles can reach similar score regions through different trade-offs among those components.

\section{Limitations}

IIQ is a proposed measurement framework, not a validated causal model. Several limitations follow directly from that status.

First, the leverage, complexity, novelty, and autonomy parameters are partly heuristic and will require empirical calibration if the metric is deployed in a specific organization. Second, title-based mappings for organizational leverage are coarse proxies for real decision scope and may fail in matrixed organizations, cross-functional teams, or public-sector hierarchies that do not align with the illustrative tiers. Third, prompt novelty is difficult to measure robustly: edit distance can miss semantic similarity, embedding similarity can over-collapse distinct but related tasks, and high-quality novelty estimation may require richer task traces than a single prompt string. Fourth, autonomy is only partially observable; active runtime, tool-use depth, and number of agent turns may each capture different aspects of agentic work. Fifth, the financial interpretation layer remains a stylized proxy even after bounding hours saved, because productivity effects may be nonlinear, delayed, or partially offset by oversight and verification costs.

A further limitation concerns organizational aggregation. Although IIQ is defined at the user level and can be aggregated to teams, departments, or the enterprise, a simple average may obscure the distribution of AI integration across the organization. For example, one highly active power user could make a team appear AI-mature even if most employees rarely use AI in their workflows. Organizational reporting should therefore include multiple aggregation views rather than a single mean score: median IIQ to reduce outlier dominance, active-user share to measure adoption breadth, top-decile share to identify concentration among power users, department-level IIQ to support internal benchmarking, IIQ concentration or Gini-style measures to distinguish broad adoption from elite-only use, and workflow-weighted IIQ to connect AI integration to critical business processes. These views would make the organizational interpretation of IIQ more robust by showing not only how much AI is used, but how broadly, evenly, and strategically it is embedded across the organization.

These limitations do not make the framework unusable, but they do constrain how it should be interpreted. The strongest near-term use case is comparative monitoring under a stable internal rubric, not high-confidence claims about absolute economic return.

\section{Conclusion}

This paper formalizes the Intelligence Impact Quotient as a structured way to measure organizational AI impact and integration beyond raw activity counts. By combining novelty-weighted token stock, distinct-task frequency, a grace-period recency gate, organizational leverage, task complexity, and autonomy, the framework distinguishes between superficial experimentation, repetitive prompting, and more meaningful deployment in consequential workflows. The normalized IIQ index and bounded interpretation layer make the metric easier to compare and communicate, while the limitations section clarifies that these outputs remain model-based rather than causal. Future work should focus on empirical calibration of novelty and autonomy signals, validation against observed workflow outcomes, and robustness checks across organizations with different role structures and task taxonomies.


\begin{thebibliography}{9}

\bibitem[Anthropic Economic Research(2026)]{anthropic-primitives}
Anthropic Economic Research.
2026.
Anthropic Economic Index: New building blocks for understanding AI use.
\url{https://www.anthropic.com/research/economic-index-primitives}.

\bibitem[Anthropic Economic Research(2025)]{anthropic-software}
Anthropic Economic Research.
2025.
Anthropic Economic Index: AI's impact on software development.
\url{https://www.anthropic.com/research/impact-software-development}.

\bibitem[McCain et~al.(2026)]{mccain2026autonomy}
Miles McCain et~al.
2026.
Measuring AI agent autonomy in practice.
Anthropic.
\url{https://www.anthropic.com/research/measuring-agent-autonomy}.

\bibitem[Kasirzadeh and Gabriel(2025)]{kasirzadeh2025characterizing}
Atoosa Kasirzadeh and Iason Gabriel.
2025.
Characterizing AI Agents for Alignment and Governance.
arXiv preprint arXiv:2504.21848.
\url{https://arxiv.org/abs/2504.21848}.

\bibitem[Kwa et~al.(2025)]{kwa2025longtasks}
Thomas Kwa, Ben West, Joel Becker, et~al.
2025.
Measuring AI Ability to Complete Long Tasks.
METR.
\url{https://metr.org/blog/2025-03-19-measuring-ai-ability-to-complete-long-tasks/}.

\bibitem[Kapoor et~al.(2024)]{kapoor2024agents}
Sayash Kapoor, Benedikt Stroebl, Zachary S. Siegel, Nitya Nadgir, and Arvind Narayanan.
2024.
AI Agents That Matter.
arXiv preprint arXiv:2407.01502.
\url{https://arxiv.org/abs/2407.01502}.

\bibitem[Humlum and Vestergaard(2025)]{humlum2025still}
Anders Humlum and Emilie Vestergaard.
2025.
Still Waters, Rapid Currents: Early Labor Market Transformation under Generative AI.
National Bureau of Economic Research Working Paper 33777.
\url{https://doi.org/10.3386/w33777}.

\bibitem[Fang et al.(2025)]{fang2025generative}
Lu Fang, Zhe Yuan, Kaifu Zhang, Dante Donati, and Miklos Sarvary.
2025.
Generative AI and Firm Productivity: Field Experiments in Online Retail.
arXiv preprint arXiv:2510.12049.
\url{https://doi.org/10.48550/arXiv.2510.12049}.

\bibitem[Patwardhan et al.(2025)]{patwardhan2025gdpval}
Tejal Patwardhan, Rachel Dias, Elizabeth Proehl, Grace Kim, Michele Wang,
Olivia Watkins, Sim{\'o}n Posada Fishman, Marwan Aljubeh, Phoebe Thacker,
Laurance Fauconnet, Natalie S. Kim, Patrick Chao, Samuel Miserendino,
Gildas Chabot, David Li, Michael Sharman, Alexandra Barr, Amelia Glaese,
and Jerry Tworek.
2025.
GDPval: Evaluating AI Model Performance on Real-World Economically Valuable Tasks.
arXiv preprint arXiv:2510.04374.
\url{https://doi.org/10.48550/arXiv.2510.04374}.

\end{thebibliography}
\end{document}